\documentclass[letterpaper, 10 pt, conference]{ieeeconf}  %

\IEEEoverridecommandlockouts                              %

\usepackage[T1]{fontenc}
\usepackage{ulem}
\usepackage{comment}
\usepackage{booktabs}
\usepackage{caption}

\usepackage{adjustbox}
\usepackage[ruled, vlined]{algorithm2e}
\usepackage{amsmath}
\usepackage{amssymb}
\usepackage{array}  %
\usepackage{authblk}
\usepackage[accsupp]{axessibility}  %
\usepackage{booktabs}
\usepackage{capt-of}
\makeatletter
\let\NAT@parse\undefined
\makeatother
\usepackage{cite}
\usepackage[font={small}]{caption}
\usepackage{colortbl}
\usepackage{epsfig}
\usepackage{inconsolata}  %
\usepackage{graphicx}
\usepackage{listings}
\usepackage{multicol}
\usepackage{multirow}
\usepackage{paralist}
\usepackage{pifont}
\usepackage{ragged2e} %
\usepackage{times}
\usepackage{tabularx}
\usepackage{xcolor}
\usepackage{svg}

\lstdefinestyle{pseudopy}{
    language=Python,
    basicstyle=\ttfamily\small,
    keywordstyle=\color{blue},
    stringstyle=\color{orange},
    commentstyle=\color{gray}\itshape,
    showstringspaces=false,
    frame=none,
    columns=fullflexible,
    keepspaces=true,
    morekeywords={function, State, struct}
}

\newcommand{\coolname}{\textit{ASP}}

\newcommand{\pifast}{\ensuremath{\pi_0}\texttt{-FAST}\xspace}
\newcommand{\pihalf}{\ensuremath{\pi_{0.5}}\xspace}

\newcommand{\object}{\ensuremath{\mathbf{o}}\xspace}
\newcommand{\objset}{\ensuremath{\mathcal{M}}\xspace}
\newcommand{\objsetglob}{\ensuremath{\mathcal{M}_{glob}}\xspace}
\newcommand{\mapfunc}{\ensuremath{\mathcal{F}_{VLM}}\xspace}
\newcommand{\objpcd}{\ensuremath{\mathcal{P}_{i}}\xspace}
\newcommand{\objfeatures}{\ensuremath{\mathbf{f}_{i}}\xspace}
\newcommand{\objrgbcrops}{\ensuremath{\mathbf{I}^{rgb}_{i}}\xspace}
\newcommand{\objdepthcrops}{\ensuremath{\mathbf{I}^{depth}_{i}}\xspace}

\newcommand{\mtt}[1]{\ensuremath{\mathtt{#1}}}

\newcommand{\goal}{\ensuremath{q_{task}}\xspace}
\newcommand{\grquery}{\ensuremath{q_{R}}\xspace}
\newcommand{\actionquery}{\ensuremath{q_{a}}\xspace}

\newcommand{\obs}{\ensuremath{X}}
\newcommand{\act}{\ensuremath{a}}
\newcommand{\actspace}{\ensuremath{\mathcal{A}}}
\newcommand{\pol}{\ensuremath{\pi}}

\newcommand{\agentact}{\ensuremath{\tilde{\act}}}
\newcommand{\agentactspace}{\ensuremath{\tilde{\actspace}}}
\newcommand{\context}{\ensuremath{c}}
\newcommand{\polagent}{\ensuremath{\pol}}
\newcommand{\polasp}{\ensuremath{\polagent_{\coolname{}}}\xspace}

\newcommand{\tool}{\ensuremath{\tau}}
\newcommand{\toolout}{\ensuremath{y}\xspace}
\newcommand{\toolset}{\ensuremath{\mathcal{T}}}
\newcommand{\su}{\ensuremath{\toolset_{G}}\xspace}
\newcommand{\skills}{\ensuremath{\mathcal{T}_\pi}\xspace}
\newcommand{\inventory}{\ensuremath{\mathcal{I}}}
\newcommand{\affmap}{\ensuremath{\mathcal{U}_{\text{VLM}}}\xspace}
\newcommand{\affpcd}{\ensuremath{\mathcal{P}_{\text{part}}}\xspace}
\newcommand{\heldobj}{\ensuremath{\mathbf{h}\xspace}}
\newcommand{\skillswitch}{\ensuremath{\tool_\pol}\xspace}
\newcommand{\retrievaltool}{\ensuremath{\tool_{R}}\xspace}
\newcommand{\navtool}{\ensuremath{\tool_{nav}}\xspace}

\definecolor{flodarkpurple}{rgb}{0.288,0.1196,0.7}

\usepackage[backref=page,breaklinks,colorlinks,bookmarks,citecolor=flodarkpurple]{hyperref}

\title{\LARGE \bf
Agentic Scene Policies:\\ Unifying Space, Semantics, and Affordances for Robot Action
}

\renewcommand{\baselinestretch}{.995}

\author{Sacha Morin$^{1,2}$, Kumaraditya Gupta$^{1,2}$, Mahtab Sandhu$^{1,2}$, Charlie Gauthier$^{1,2}$,\\
Francesco Argenziano$^{3}$, Kirsty Ellis$^{1,2}$, and Liam Paull$^{1,2}$%
\thanks{*The work at the Université de Montréal was supported by the Natural Sciences and Engineering Research Council of Canada (NSERC) (Paull) and an NSERC PGS D Scholarship (Morin).}
\thanks{$^{1}$Université de Montréal}%
\thanks{$^{2}$Mila -- Quebec AI Institute}%
\thanks{$^{3}$Sapienza University of Rome}%
}

\begin{document}
\bstctlcite{IEEEexample:BSTcontrol} %

\makeatletter
\let\@oldmaketitle\@maketitle
\renewcommand{\@maketitle}{\@oldmaketitle
\centering
\includegraphics[width=0.87\textwidth]{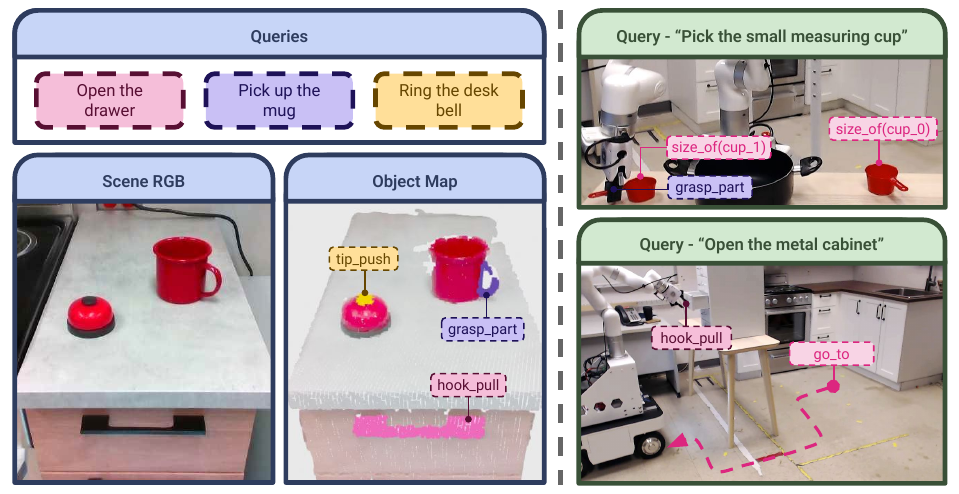}
\captionof{figure}{
\textbf{\coolname{}} is an agentic system for zero-shot language-based tabletop and mobile manipulation. Core to our approach is a scene-agent tool interface that blends object retrieval, spatial reasoning, and affordance-based skills (e.g., \texttt{grasp\_part}, \texttt{tip\_push}) enabling interactions with particular parts (e.g., \texttt{handle}, \texttt{button}) across a large variety of objects.
(\href{https://montrealrobotics.ca/agentic-scene-policies.github.io/}{Webpage})
}

\label{fig:splash}
\vspace{-0.98em}
}
\makeatother

\maketitle

\thispagestyle{empty}
\pagestyle{empty}

\begin{abstract}
Designing or learning robot policies that generalize zero-shot across a range of language instructions and objects is a core problem in robotics. Vision-Language-Action models~(VLAs) learn such policies end-to-end by repurposing existing Vision-Language Models~(VLMs), but generalization to new instructions and objects remains challenging. An alternative is to implement a modular policy by leveraging an explicit VLM-based 3D scene representation and motion planning. While modular policies show strong zero-shot potential, they typically retrieve objects based on semantics without explicit spatial reasoning, severely restricting their overall grounding capabilities. They also interact with objects using basic grasping and navigation skills. In this work, we address these limitations by unifying grounding capabilities and robot skills in a single agentic action space through a scene-agent tool interface. By leveraging part-level affordances, our skills generalize across diverse objects and enable zero-shot interactions such as unplugging chargers and opening drawers. We name the resulting framework Agentic Scene Policies~(\coolname{}). Through extensive real-world experiments, we show how \coolname{} consistently outperforms leading VLAs in the zero-shot setting. We also demonstrate the extensibility of our framework by introducing a mobile version of \coolname{} to tackle room-level queries. See our \href{https://montrealrobotics.ca/agentic-scene-policies.github.io/}{project page} for more results.

\end{abstract}

\setcounter{figure}{1} %

\section{INTRODUCTION}

Language-conditioned robot policies need to manage the complex interplay between language, space, and action. Much of the recent progress on this problem has been driven by vision-language models (VLMs) trained on internet-scale data and showing strong general visual understanding in the open-world. Recent methods use VLMs as backbones for end-to-end policy learning, yielding Vision-Language-Action models (VLAs)~\cite{ 
zitkovich2023rt2,  
pertsch2025fast, intelligence2025pihalf 
} that directly map sensor data and language commands to robot actions. VLAs aim to leverage the broad multimodal reasoning and instruction following capabilities of VLMs to achieve more general robot control. However, there is accumulating evidence that VLAs can struggle with semantic generalization~\cite{gao2026taxonomy} and diminish the capabilities of the underlying VLM during fine-tuning~\cite{hancock2025actions}. %

There is therefore substantial interest in using VLMs \textit{as is} for robot planning and control. This can be achieved by using VLMs to build explicit open-vocabulary 3D scene representations. Combining language-based object retrieval in these representations with motion planning has yielded capable modular systems for zero-shot language-conditioned mobile manipulation~\cite{vlmaps, liu2024ok, gu2024conceptgraphs, liu2025dynamem}. We refer to such systems as \textit{Scene-based Policies}.

While they preserve the generality of VLMs, scene-based policies face some key limitations and cannot act as drop-in replacements for VLAs. First, they are limited in terms of \uline{grounding} capabilities, often implementing ``argmax'' queries where they simply interact with the object that is most similar with the current language query, with limited contextual understanding of the scene or the task. Second, they are limited in terms of \uline{physical interactions} and are often constrained to navigation and grasping. 

In this work, we address both limitations by modeling a scene-based policy as an LLM agent. We propose a \emph{tool-centric} scene interface to unify object grounding and robot execution capabilities. We define skills at the level of \emph{affordances}, allowing the agent to reuse them across a variety of objects and language instructions in a zero-shot manner. We name the resulting framework Agentic Scene Policies~(\coolname{}). Our key contributions are

\begin{enumerate}
\item We introduce \coolname{}, a scene-based agentic policy managing a symbolic inventory of objects for spatial grounding and affordance-based skill execution (Section~\ref{sec:method}).
\item We find that the orchestration of the \coolname{} grounding and skills tools consistently outperforms leading VLAs (\pifast, \pihalf) in the zero-shot setting over a range of objects and commands (Section~\ref{sec:results}). Our experiments provide a valuable data
point in the ongoing debate between modular and end-to-end methods.
\item We introduce a mobile version of \coolname{}, demonstrating the extensibility of our framework (Section~\ref{subsec:mobile}), and evaluate it against a policy based on \textit{ConceptGraphs}~\cite{gu2024conceptgraphs}.
\end{enumerate}

\section{RELATED WORK}

\textbf{Vision-Language-Action Models.} VLAs learn policy functions end-to-end through imitation learning using an architecture integrating a pretrained VLM as a backbone~\cite{
zitkovich2023rt2,
black2024pi_0,   
pertsch2025fast, intelligence2025pihalf}. The web-scale pretraining of the VLM component broadens the range of visual and linguistic concepts supported by the policy, enabling semantic generalization.
While some zero-shot potential has been observed, VLAs often require some fine-tuning on specific problems to perform well~\cite{kim2025fine}. 

\textbf{Open-Vocabulary Mapping.}  %
Open-vocabulary maps replace conventional class labels in semantic maps with multi-modal features from foundation models such as CLIP
~\cite{gu2024conceptgraphs, 
slam-handbook, kassab2025openlex3d, koch2024open3dsg, conceptfusion, peng2023openscene, 
hovsg, lu2023ovir}, allowing a much richer representation of semantics. For mapping, features are typically extracted from vision sensors, grounded to a map element (point, voxel, object), and aggregated across views. At inference time, they are compared with query features, enabling specific queries about objects and their properties. Open-vocabulary maps underpin recent scene-based systems for navigation~\cite{vlmaps} and mobile manipulation~\cite{liu2024ok, liu2025dynamem, gu2024conceptgraphs}.

\textbf{Affordance Detection.}  Affordance detection aims to segment functional elements of objects, such as handles, buttons, or knobs, that enable specific interactions~\cite{
delitzas2024scenefun3d,
Winston1983LearningPD,
affordancenet,Luddecke_2017_ICCV, Nagarajan2020LearningAL
}. They have the potential to greatly simplify the implementation of robot skills by providing direct cues about the geometry that a robot can manipulate as well as the manner in which it should be manipulated. Recent methods have focused on leveraging VLMs for this task~\cite{openfungraph, fun3du,
tong2024oval, qian2024affordancellm
} given their increasingly good performance at detecting parts in images~\cite{team2023gemini}. 

\textbf{LLM Agents for Robotics.} Taking advantage of the coding abilities of LLMs, recent work has framed language-conditioned robot policies as a translation problem between natural language and code, effectively mapping language commands to robot API calls, LLM-generated function calls, or skills~\cite{code-as-policies, singh2022progprompt, saycan}, with various strategies existing for scene grounding~\cite{voxposer, gu2024conceptgraphs}. This has been followed by more reactive \underline{agentic} approaches where LLMs interleave text generation and tool calls instead of coding entire plans upfront~\cite{yao2023react, robot-agent-survey, llmgrounder}, enabling a tighter feedback loop with the API and other frameworks such as ROS~\cite{rosa}. 

Our work can be understood as proposing an agentic harness for a base LLM model to implement zero-shot tabletop and mobile manipulation policies. While \coolname{} leverages off-the-shelf components (e.g., ConceptGraphs for 3D mapping~\cite{gu2024conceptgraphs}, AnyGrasp for grasp prediction~\cite{fang2023anygrasp}), our core contribution is a stateful tool interface that blends spatial reasoning and affordance-driven skill execution. Concurrent work also explores harnesses for agentic robotics and evaluate the resulting systems against end-to-end VLAs~\cite{guo2026agenticlab, chen2026gap}. More generally, designing action spaces and memory modules for LLM agents plays a central role in unlocking the full capabilities of the base models~\cite{coala}.

\section{PRELIMINARIES}
\label{sec:preliminaries}

\textbf{Language-conditioned Policies.} We consider the general problem of mapping a language instruction $\goal$ and the current observation $\obs_t$ to a robot action $\act_t \in \actspace$ with a policy $\pol(\act_t|\obs_t, \goal)$. 
For simplicity, the following sections focus on the tabletop setting. Section~\ref{sec:implementation} will cover mobile manipulation and multiple input keyframes.

\textbf{Object Map.} We  assume access to a VLM-based open-vocabulary mapping function $\mapfunc$ that maps $\obs_t$ to a set of 3D objects $\objset_t = \mapfunc(\obs_t) = \{\object_i\}_{i=1}^k$. We define each object as a tuple $\object_i = \langle \objpcd, \objfeatures, \objrgbcrops, \objdepthcrops \rangle$ representing respectively the object 3D point cloud, the object-centric CLIP features~\cite{clip}, and the RGB-D object crops. Our object definition is similar to \cite{gu2024conceptgraphs} and we will detail \mapfunc in Section~\ref{sec:implementation}.

\textbf{Scene-based Policies.} We  model a scene-based policy  $\pol(\act_t|\objset_t, \goal)$. Scene-based policy functions are typically \uline{not learned} and instead combine language-based object retrieval and classical motion planning in a modular system to implement behaviors. 

\textbf{Agents as Policies.}  $\text{LLM}$ agents can interact with the world by calling tools $\tool^i$ from a set $\toolset=\{\tool^i\}_{i=1}^N$. Tools map text arguments $\theta$ to some text output $\toolout$, i.e. $\tool^i(\theta) = \toolout$, and give the agent access to external resources, typically for information retrieval (databases, search engines), computation (code interpreters), and action (emails, robot skills).

Agents can be interpreted as policies by considering an agentic action space $\agentactspace$ where actions are all possible tuples $\agentact =\langle \tool^i, \theta\rangle$ representing a choice of tool and corresponding tool arguments. A given action $\agentact_t=\langle \tool^i_t, \theta_t\rangle$ can be executed to yield an output $\toolout_t=\tau^i_t(\theta_t)$. The agent recursively builds a context $\context_t=(\agentact_0, \toolout_0, \agentact_1, \toolout_1,\dots, \agentact_{t-1}, \toolout_{t-1})$ by sampling a policy $\polagent(\agentact_t |\goal, \context_t)=\text{LLM}(\toolset, \goal, \context_t)$ with context transitions $\context_{t+1} = (\context_t, \agentact_t, \toolout_t)$. A particular episode ends when the agent stops calling tools.

In the following sections, we show how the agentic action space $\agentactspace$ can act as a unified interface combining object grounding capabilities in $\objset_t$ with affordance-based skills triggering robot actions in $\actspace$, yielding an effective zero-shot scene-based manipulation policy. We achieve this by imposing a particular structure on the tool set $\toolset$.

\begin{figure}[!ht]
    \centering
    \includegraphics[width=\linewidth]{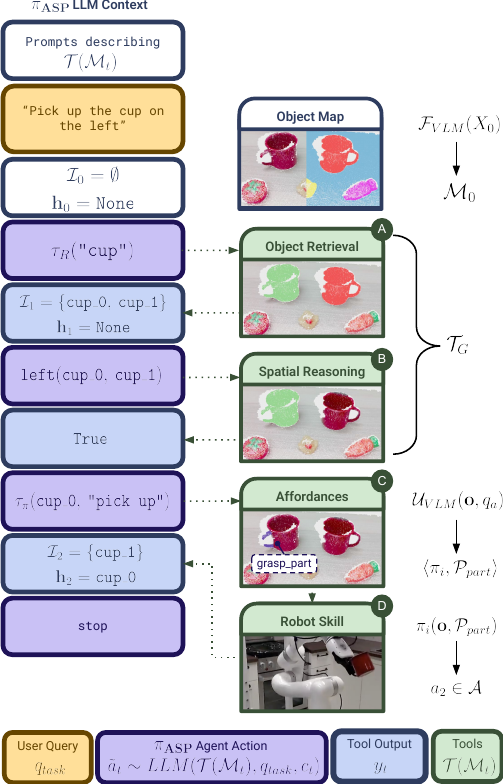}
    \caption{
    \textbf{Framework \& Illustrative Query.} We illustrate \coolname{} on the query $\goal=$\texttt{"Pick up the cup on the left"}. 
    Given the RGB-D observation $\obs_0$, we first build an open-vocabulary object map $\objset_0 = \mapfunc(\obs_0)$. 
  The agent \polasp performs actions $\agentact_t \in \agentactspace$~(Section~\ref{sec:preliminaries}) by calling tools $\tool \in \toolset(\objset_t)$, and expanding the context $\context_t$ with tool outputs $\toolout_t$. 
  Tools update a symbolic state $\langle \inventory_t, \heldobj_t \rangle$ encoding the current inventory of grounded objects and the held object.  \textbf{(A)} The agent retrieves task-relevant objects using
  $\retrievaltool$, adding two cups to the inventory $\inventory_t$. \textbf{(B)} It disambiguates the correct cup using a \mtt{left}  spatial reasoning tool, and then \textbf{(C)} sends the \texttt{"pick up"} action description to the skill switch
  $\skillswitch$, which internally runs the affordance predictor $\affmap(\object, \actionquery)$ to select a skill $\pol_i \in \skills$ (\mtt{grasp\_part}) and segment the
  relevant point cloud $\affpcd$ (handle). \textbf{(D)} The selected skill $\pol_i(\object, \affpcd)$ is executed on the robot, updating the state to reflect the grasped
  object ($\heldobj_2 = \texttt{cup\_0}$). The specific sequence of tool calls is determined by the agent and will vary across queries; for example, simpler
   instructions may skip spatial reasoning, while others may require multiple retrieval calls or different skills.
    }
    \vspace{-5.0mm}
    \label{fig:pipeline}
\end{figure}

\section{METHOD}
\label{sec:method}
\textbf{Overview and Example Query.} Agentic Scene Policies~(\coolname{}) models a scene-based policy by conditioning the tool set of an LLM agent on the object set $\objset_t$, which we denote as $\polasp(\agentact_t | \objset_t, \goal, \context_t)=\text{LLM}(\toolset({\objset_t}), \goal, \context_t)$. 

We will introduce tools using an example query $\goal=$\texttt{"Pick up the cup on the left"} in a scene with distractors (Fig.~\ref{fig:pipeline}). The tool set $\toolset({\objset_t})$ first includes a set of \uline{grounding tools} \su for object retrieval (\textit{Which objects in $\objset_t$ are cups?}) and spatial reasoning (\textit{Which cup is on the left?}). Once the task-relevant objects are grounded, \polasp can employ \uline{affordance-based skill tools} \skills available through a language-conditioned skill switch $\skillswitch$ (\textit{Which skill achieves the "pick up" action on the cup?}). We design $\toolset(\objset_t)$ such that \polasp can infer the correct order of grounding and skill tool calls to solve a large number of manipulation queries. We will provide implementation details for each tool in Section~\ref{sec:implementation}.

\textbf{Agent State Variables.} Tools return scene information to \polasp using a symbolic state tuple $\langle \inventory_{t}, \heldobj_t \rangle$. The set $\inventory_{t} \subseteq \objset_t$ represents an inventory of task-relevant objects that are retrieved from the scene and can be processed by other tools, while the variable $\heldobj_t$ describes the currently held object (if any). We initialize $\inventory_0=\emptyset$ and $\heldobj_0=\mathtt{None}$. %

\textbf{Grounding Tools.} The agent \polasp will first rely on the grounding tools $\su$ to understand the scene context and ground the objects that are relevant to the query $\goal$.  Tools in $\su$ are only used for \uline{reasoning}. They expand the current agent context $\context_t$ using information from $\objset_t$, without actuating the robot.

To identify task-relevant objects, the agent $\polasp$ will iteratively grow the inventory $\inventory_t$ using a \uline{retrieval tool} $\retrievaltool(\grquery)=\inventory_{t+1}=\inventory_{t} \cup \objset_t(\grquery)$ where $\grquery$ is a text query and $\objset_t(\grquery) \subseteq \objset_t$ is the subset of objects that match $\grquery$. We emphasize that (i) $\objset_t(\grquery)$ may include multiple relevant objects (Fig.~\ref{fig:pipeline}-A, the two cups in the scene), and (ii) $\polasp$ may choose to make multiple calls to \retrievaltool to ground distinct objects that are relevant to $\goal$ (e.g., make a separate call for \texttt{"plate"} if we need to place the left cup on a plate).

Language-conditioned policies are expected to understand multi-object referencing and need to differentiate inventory objects based on relative positions or properties. We therefore extend $\su$ with \uline{spatial reasoning tools} that can measure quantities (e.g., distances) or check predicates in between objects in $\inventory_t$. For example, a left of predicate tool would take the form $\mathtt{left}(\object_1, \object_2): \inventory_t \times \inventory_t \mapsto \{\mathtt{True, False}\}$ and return a boolean based on the available geometry in $\objset_t$ (Fig.~\ref{fig:pipeline}-B). The combination of retrieval and spatial reasoning tools in $\su$ will allow \polasp to ground an arbitrary number of task-relevant objects in the scene. Spatial reasoning tools return primitive types (e.g., \texttt{float}, \texttt{bool}) without changing  $\inventory_{t}$ and $\heldobj_{t}$.

\textbf{Skill Switch Tool.} Once a relevant object has been identified, \polasp can act on it using a set of object-centric skills $\skills=\{\pol_i\}_{i=1}^k$, which includes a range of part-level interactions such as grasping, pushing, and pulling. $\skills$ is not directly available to \polasp. Instead, we expose a language-conditioned skill switch $\skillswitch(\object, \actionquery)$ where \polasp specifies $\object \in \inventory_t$ and the intended action $\actionquery$ in text form (e.g., \texttt{pick up}, \texttt{open}, \texttt{unplug}). Internally, \skillswitch first runs a \uline{VLM-based affordance predictor} $\affmap(\object, \actionquery)=\langle \pol_i, \affpcd \rangle$, representing an affordance as the combination of a skill $\pi_i \in \skills$ and the relevant 3D part-level point cloud $\affpcd$ (Fig.~\ref{fig:pipeline}-C). The skill switch $\skillswitch$ then calls the selected skill $\pol_i(\object, \affpcd)$. 

\textbf{Skill Tools.} A skill $\pol_i$ computes a robot action $\act_t \in \actspace$ using motion planning and \affpcd. In practice, our skills output sequences of end-effector positions~(Section~\ref{subsec:results_tool_details}). Skills then run $a_t$ on the robot and return the updated state variables $\langle \inventory_{t+1}$, $\heldobj_{t+1} \rangle$ to the agent. For example, a picking tool may remove an object from $\inventory_{t+1}$ and add it to held object variable $\heldobj_{t+1}$ (Fig.~\ref{fig:pipeline}-D) while a placing tool may set $\heldobj_{t+1}$ to \texttt{None}. 

\textbf{Action Design.} The design of $\skillswitch$ and $\skills$ is guided by three primary considerations. First, affordance-based skills are specific enough to have tractable implementations, and general enough to be reused zero-shot across multiple objects. We hypothesize that the composition of $\su$ and the affordance-based $\skills$ will enable interactions with a broad variety of objects and outperform end-to-end learning in the zero-shot setting. Second, \affmap allows skill selection to be informed by vision (given $\objrgbcrops$ in $\object_i$, see Section~\ref{subsec:results_tool_details}) and influenced by the existence of a part (e.g., not all cups have handles) and their shapes (e.g., to try hooking or pulling depending on the handle shape). Third, we only compute affordances when the agent requests an object interaction, avoiding the perception overhead of running affordance detection on all objects in $\objset_t$ which can be high when using general skills with ``dense'' support over many objects. For example, any small object can be potentially grasped, pushed or pulled.

\textbf{Tool Feedback.} Tools can verify some pre-conditions. Examples include checking if $\object \in \inventory_t$ when calling a spatial reasoning tool or ensuring that the held object $\heldobj_t=\mathtt{None}$ when grasping. Tools validate these conditions explicitly and may return a feedback string to $\polasp$ in addition to their usual outputs, allowing $\polasp$ to reattempt tool calls.

\textbf{Remapping.} Skills in \skills may verify post-conditions (e.g., is the gripper holding something). In case of failure, they may trigger \uline{remapping} and return an empty inventory $\inventory_{t+1}=\emptyset$, signaling to $\polasp$ that regrounding will be necessary. In this case, we refresh the object set $\objset_{t+1}=\mapfunc(X_{t+1})$ to account for possible object movements caused by a manipulation failure. Absent a remapping trigger, we simply set $\objset_{t+1}=\objset_t$ to minimize perception overhead.

\textbf{Object Identifiers.} For simplicity of exposition, tools in this section use objects or object sets as inputs and outputs. In practice however, the $\polasp$ agent only interacts with unique string identifiers generated by the grounding tool $\retrievaltool$ (e.g., \texttt{cup\_0}), as opposed to a full serialization of the object content. Tools internally use the identifier to retrieve the relevant object from $\objset_t$.

\section{IMPLEMENTATION DETAILS}
\label{sec:implementation}
In this section, we present a particular instantiation of \coolname{} following the scaffolding laid out in Section~\ref{sec:method}. While we define particular sets of grounding tools $\su$ and skill tools \skills, we note that they can easily be extended and even potentially generated~\cite{code-as-policies, singh2022progprompt}. Agent code (including specific hyperparameters and prompts) is available \href{https://github.com/montrealrobotics/agentic-scene-policies}{here}.

\subsection{Object Map}
\label{subsec:results_map_implementation}
\textbf{Open-Vocabulary Object Map.} We define \mapfunc similarly to recent work on open-vocabulary object-centric map representations~\cite{gu2024conceptgraphs, lu2023ovir}. Whenever we run \mapfunc in the tabletop setting, we position the arm at a reference ``home'' pose to have a good overview of the workspace and process the RGB-D wrist frame as $\obs_t$. We first run class-agnostic segmentation (MobileSAM~\cite{mobilesam, sam} in grid-sampling mode) to extract segmentation masks and convert them to bounding boxes. For each mask, we crop a local RGB image $\objrgbcrops$ and embed it with CLIP to describe semantics, yielding $\objfeatures$. We also backproject the masks using the camera depth ($\objdepthcrops$) to obtain the 3D point clouds $\objpcd$.

\subsection{Agent \& Tools}
\label{subsec:results_tool_details}

\textbf{Agent.} We implement the agent \polasp using \href{https://github.com/langchain-ai/langchain}{LangChain} with the package's default prompts and a Gemini backend~\cite{team2023gemini}. The agent is prompted with descriptions of $\inventory_t$, $\heldobj_t$ and all the tools in \su and \skills.

\textbf{Object Retrieval.} We implement $\retrievaltool(\grquery)$ as a search for the top $k$ similar objects in $\objset_t$ based on the similarity between $\text{CLIP}(\grquery)$ and the object features $\objfeatures$ in $\objset_t$. We then use a VLM to confirm whether each of the top $k$ objects are relevant to the $\grquery$ or not using the crops $\objrgbcrops$~\cite{liu2025dynamem}. %
In our experiments, we use $k=3$ and Gemini 2.5 as the VLM classifier. 

\textbf{Spatial Reasoning Tools.} We expose \mtt{distance\_to}$(\object)$, \mtt{distance\_between}$(\object_1, \object_2)$, \mtt{left}$(\object_1, \object_2)$, \mtt{right}$(\object_1, \object_2)$, and \mtt{size\_of}$(\object)$ tools. Spatial reasoning tools are implemented as basic operations on the object point clouds $\objpcd$.

\textbf{Affordance Detection.} The affordance extractor \affmap$(\object_i, \actionquery)$ implements a two-step process. (Step 1) We first use Gemini 2.5 to predict a $\langle\pol, q_{part}\rangle$ tuple given the RGB crop $\objrgbcrops$, with $\pol \in \skills$ and $q_{part}$ describing the relevant part in language. (Step 2) Again with Gemini 2.5, we predict a bounding box given $\objrgbcrops$ and $q_{part}$, process them with SAM 2.1 to obtain masks, and lift them to 3D using \objdepthcrops to yield \affpcd. 
We derived this approach after observing that VLMs are better at detecting parts given the explicit part description $q_{part}$ instead of the action description $\actionquery$. We skip Step 2 for general object skills that do not require \affpcd.

\textbf{Skill Tools.} We implement all skills in \skills using motion planning. We first expose some general object skills:

\begin{itemize}
    \item $\mtt{grasp}(\object)$: grasp \object anywhere.
    \item $\mtt{place}(\object)$:  place held object $\heldobj_t$ on \object.
    \item $\mtt{drop}(\object)$:  drop held object $\heldobj_t$ on \object.
\end{itemize}

as well as the following affordance-based skills, many of which are inspired by the SceneFun3D affordance types~\cite{delitzas2024scenefun3d}:
\begin{itemize}
    \item $\mtt{grasp\_part}(\object, \affpcd)$: grasp \affpcd.
    \item $\mtt{tip\_push}(\object, \affpcd)$: push on \affpcd with the gripper tip.
    \item $\mtt{pinch\_pull}(\object, \affpcd)$: pinch \affpcd and pull.
    \item $\mtt{hook\_pull}(\object, \affpcd)$: hook \affpcd from above and pull.
\end{itemize}

Skill implementations generate sequences of end-effector pose goals for the motion planner using a combination of operations on the object point cloud $\objpcd$ (e.g., finding the normal or a point above it) and AnyGrasp~\cite{fang2023anygrasp}. In the case of part-level skills with affordances, we derive goals using \affpcd. We always run AnyGrasp on a context around the object and only keep the grasps that are near \objpcd (for $\mathtt{grasp}$) or \affpcd (for $\mathtt{grasp\_part}$). For the two grasping skills, we iterate over the top 10 grasps until we find one where the motion planner succeeds (and return an error message otherwise).  In the case of $\mathtt{pinch\_pull}$ and $\mathtt{hook\_pull}$, we use a predefined end-effector rotation to grasp the \affpcd centroid and infer a horizontal pulling axis using the local point cloud normal. For $\mathtt{grasp}$, $\mathtt{grasp\_part}$ and $\mathtt{pinch\_pull}$, we explicitly verify that the gripper holds something after execution and inform the agent of a failure if not.

\subsection{Mobile \coolname{}}
\label{subsec:mobile}
\textbf{Mobile \coolname{}.} While we primarily focused on the tabletop setting so far, we briefly discuss how \coolname{} can be extended to the mobile setting to solve room-level queries. We note that only minor variations are required given the extensible nature of $\objset_t$ and the LLM agent introduced in Section~\ref{sec:method}.

\textbf{Object Map.} In the mobile setting, we teleop the robot to collect 1-5 posed RGB-D frames covering relevant objects across the room.  We then process each frame with \mapfunc and merge objects across frames to build a global object map $\objsetglob$, identical to \cite{gu2024conceptgraphs}. We keep track of the best RGB-D crops for each object using segment area. This mapping phase takes less than one minute per query.  

\textbf{Navigation Tool.} The Mobile \coolname{} agent replaces the skill switch $\skillswitch$ with a navigation tool $\navtool(\object, \actionquery)$ to visit objects in $\objsetglob$. It otherwise follows the framework presented in Section~\ref{sec:method} and can use $\su$ to ground task-relevant objects in $\objsetglob$. Similar to the skill switch \skillswitch, the agent will pass an action description \actionquery to $\navtool$, allowing it to detect an affordance $\langle\pol_i, \affpcd\rangle$ on the target object with $\affmap$ \uline{before} starting navigation. In the mobile setting, \affmap runs affordance detection over the top RGB crops and performs multi-view aggregation, similar to how \cite{gu2024conceptgraphs} merges objects across views. Knowing the affordance orientation in advance will help navigate to a pose that is suitable for interaction.

Specifically, we infer a normal from the \affpcd point cloud (if any) oriented away from the object centroid. We project this normal on the horizontal plane to infer a preferred viewing position $\mathbf{p}_{aff} \in \mathbb{R}^2$ at a distance of $r$ from the object centroid. While pre-collecting keyframes in the mobile setting, we also build a 2D occupancy map using the navigation stack. For the exact navigation goal, we find a target pose that is in free space and close to $\mathbf{p}_{aff}$. Specifically, we search for candidate poses $(\mathbf{p} \in \mathbb{R}^2, \phi \in [0, 2\pi))$ at discrete intervals on the circle of radius $r$ centered at the object centroid by minimizing $\mathcal{C}(\mathbf{p}, \phi) + \lambda_{aff} \| \mathbf{p} - \mathbf{p_{aff}}\|_2$ with $\mathcal{C}$ denoting the average footprint cost in the occupancy map, and $\lambda_{aff}$ being a penalty factor. We always choose candidate $\phi$ to face the object and discard candidate poses with high footprint cost (collisions). We start with a small $r$ (0.85m) and progressively relax it if no valid pose is found.

\textbf{Manipulation.} Following a successful $\tool_{nav}(\object, \actionquery)$ call, we build a separate local $\objset_t$ from the current camera frame, retrieve $\object$ in $\objset_t$, and then call the skill switch $\skillswitch(\object, \actionquery)$ to interact with \object. This ``redetection'' renders \polasp more robust to possible localization errors and mapping errors in $\objsetglob$.

\subsection{Robot}
\label{subsec:results_robot}

\textbf{Hardware.} Our mobile manipulator consists of a UFactory XArm6 mounted on an Agilex Ranger Mini 2.0, similar to the build in \cite{xiong2024adaptive
}.  We mounted an Intel Realsense D435i on the arm wrist and an Intel Realsense T265 tracking camera on the base. We stream RGB-D data over wifi and run all perception models on a workstation with an NVIDIA Titan RTX.

\textbf{Software.} The robot software is integrated with ROS 2. We use MoveIt 2~\cite{chitta2012moveit} for the arm motion planning (RRT-Connect Planner).
SLAM and navigation are handled by RTABMap~\cite{rtabmap} and Nav2~\cite{nav2} using the wrist RGB-D camera and odometry estimates from the T265 and the base.

\begin{figure}
    \centering
    \includegraphics[width=\columnwidth,trim={0 0 0 0 },clip]{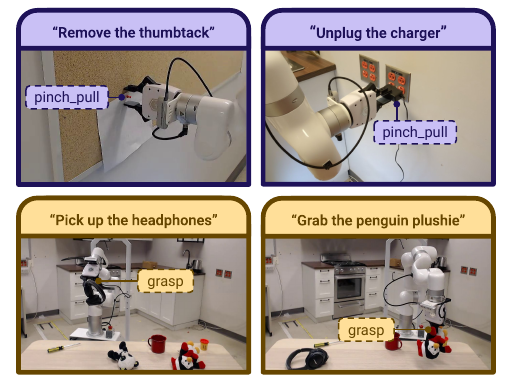}
    \caption{Examples of manipulation queries. \textbf{(Top)} \coolname{} can detect the \mtt{pinch\_pull} affordance and deploy the corresponding skill in different situations involving different verbs (\texttt{Remove}, \texttt{Unplug}), and objects. \textbf{(Bottom)} Examples of double pick queries in the mobile setting involving a variety of objects, including headphones and a penguin plushie.
    }
    \vspace{-5.0mm}
    \label{fig:cool_demos}
\end{figure}

\begin{figure*}
    \centering
    \includegraphics[width=1.0\textwidth]{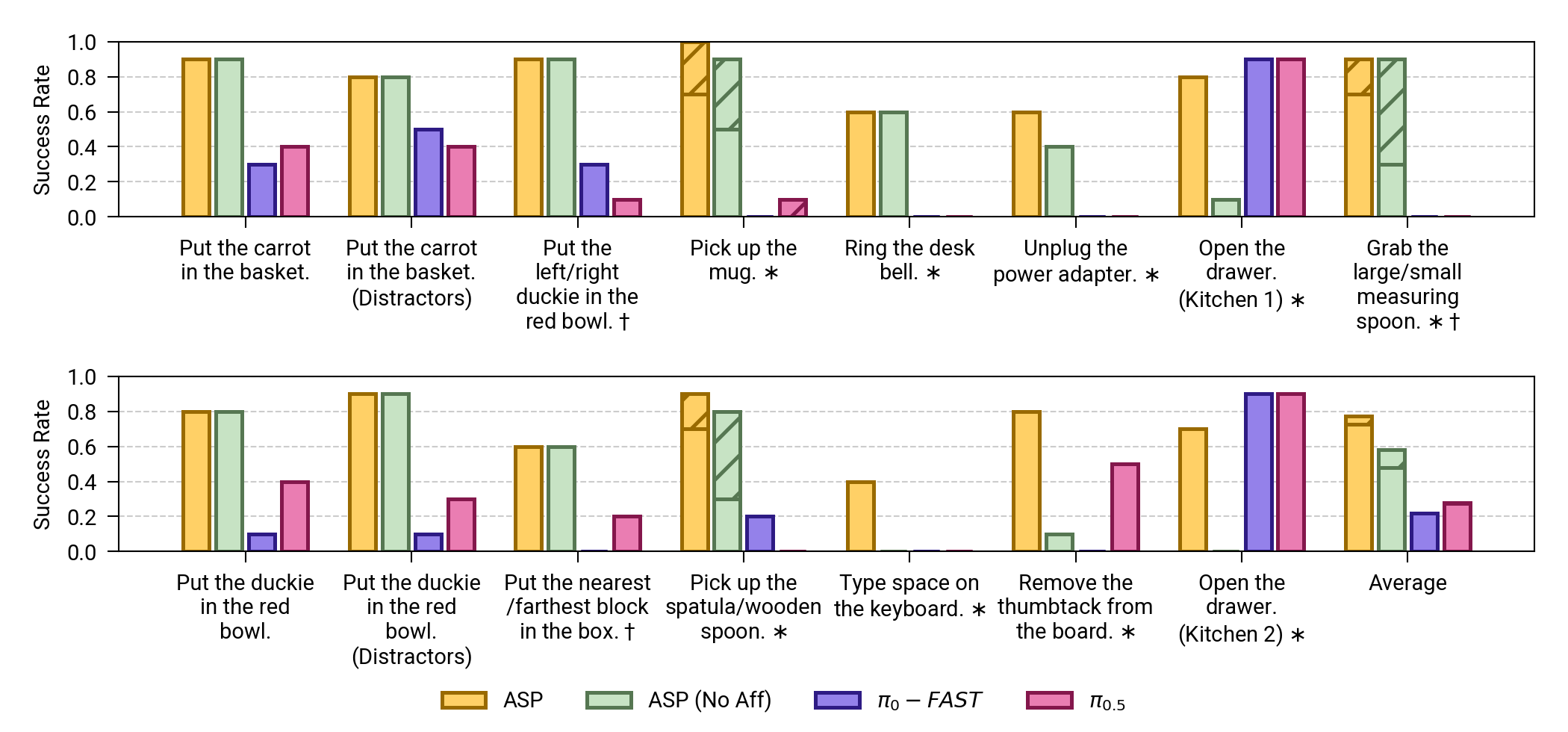}
    \caption{\textbf{Tabletop Manipulation Results.} We run a total of 540 trials to compare \coolname{} with an affordance-free ablative baseline, \coolname{} (No Aff), and two VLA baselines on 15 tabletop manipulation queries involving a variety of objects, geometric concepts (left/right, large/small), and affordances.
    \textbf{(Query Types)} Four queries are pick-and-place tasks, eight require part-level affordance interactions (marked with $\ast$), two require spatial reasoning ($\dagger$), and one requires both ($\ast\dagger$).
    \textbf{(Success Rate)} We report the average success rate over 10 attempts for each task. 
    \textbf{(Hatch Pattern)} For objects with handles, we use a hatch pattern for trials where methods succeed without using the handle or by using the handle in an unnatural way according to human judgment. 
    \textbf{(Progress Rate)} While we focus on hard success rate in this figure, we also provide some task progression numbers in Fig.~\ref{fig:progress_rate_results}.
    Videos, including failures, are available on our \href{https://montrealrobotics.ca/agentic-scene-policies.github.io/}{project page}.
    }
    \label{fig:tabletop_results}
    \vspace{-6.0mm}
\end{figure*}

\begin{figure}
    \centering
    \includegraphics[width=1.00\columnwidth]{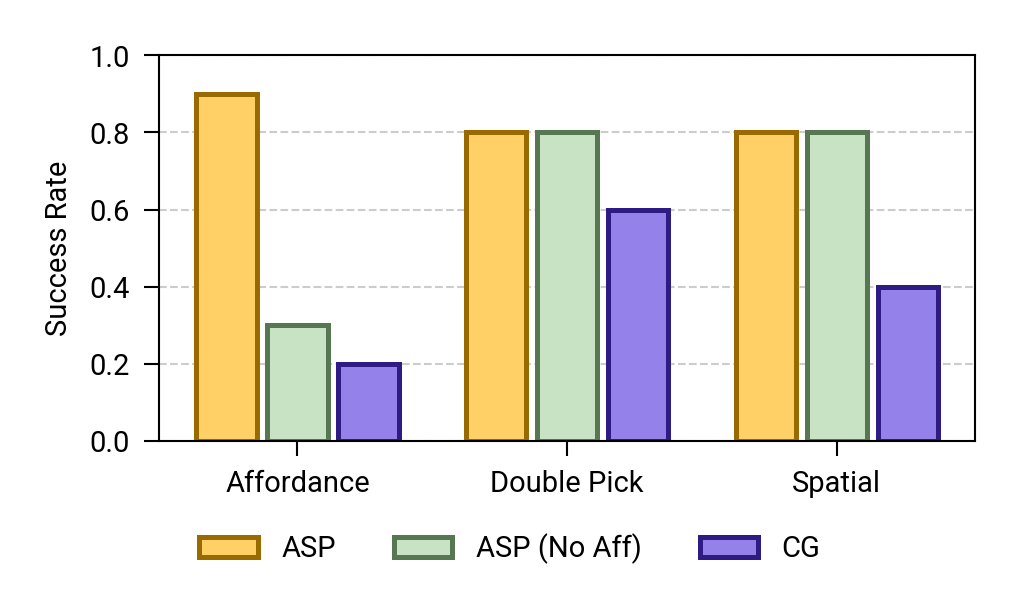}
    \caption{\textbf{Mobile Manipulation Results.} We run mobile \coolname{}, mobile \coolname{} (No Aff) and a \textit{ConceptGraphs}-based agent (CG) on queries in a medium-sized room. For each query, we teleop the robot to collect 1 to 5 keyframes from the room before launching the policy. \textbf{(Affordance)} We run two queries: \texttt{Dial a number on the phone} (5 times) and \texttt{Open the metal cabinet} (5 times) (Fig.~\ref{fig:splash}, bottom right). In both cases, the object is accessible from multiple angles but some viewing angles are better for interaction (i.e., facing the keypad and drawer) and the robot must navigate accordingly. \textbf{(Double Pick)} \coolname{} must put two objects in a target container, e.g. \texttt{Put the screwdriver and the red mug in the cardboard box}. The queries cover a large variety of objects including headphones, a penguin plushie, and a variety of toy food items (Fig.~\ref{fig:cool_demos}). We give a score of 0.5 per object in the container and no other partial points. We run 10 unique double pick queries with distractors. \textbf{(Spatial)} \coolname{} must disambiguate a mobile pick-and-place query using spatial reasoning, e.g. \texttt{Place the egg that is near the tomato in the pan}. We run 10 unique spatial queries with distractors. Methods are required to interact with 22 unique objects to solve queries in this section.} 
    \vspace{-5.0mm}
    \label{fig:mobile_results}
\end{figure}

\begin{figure*}
    \centering
    \includegraphics[width=0.90\textwidth,trim={0 0 0 0},clip]{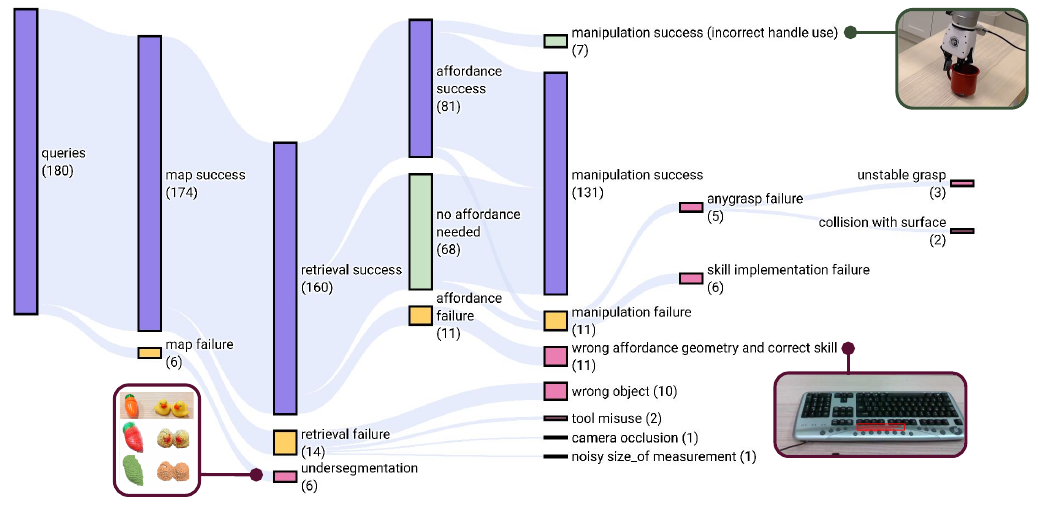}
    \caption{\textbf{Failure Analysis.} The modular nature of \coolname{} enables an in-depth analysis of failure modes across four stages: mapping (\mapfunc), object retrieval (\retrievaltool), affordance detection (\affmap), and skill execution ($\pol_i$).
    We find that a high number of failures can be attributed to perception (31 queries), highlighting potential areas of improvements in the current stack. \textbf{(Undersegmentation)} Our SAM-only \mapfunc requires tuning some hyperparameters~\cite{gu2024conceptgraphs} to merge segments when building $\objset_t$. Finding a set of hyperparameters that is optimal across queries and robust to depth noise is challenging. \textbf{(Wrong Affordance Detection.)} While \affmap is adept at mapping query verbs to the correct skill (e.g., \texttt{Type} to \mtt{tip\_push}), the detection stage is imperfect. For example, we find that Gemini can be influenced by the ``canonical'' location of a space bar when faced with an upside down keyboard. \textbf{(Incorrect Handle Use)} An example of a ``hatched'' success in Fig.~\ref{fig:tabletop_results}. While the handle was correctly segmented and \coolname{} successfully picked the mug, the grasp does not correspond to a typical handle use, showcasing the limits of segmentation-driven grasp selection. \textbf{(VLAs)} We provide a similar plot for \pifast and \pihalf in Fig.~\ref{fig:sankey_pi}.
    }
    \vspace{-6.0mm}
    \label{fig:sankey}
\end{figure*}

\section{EXPERIMENTS}
\label{sec:results}
We design our experiments around the following questions:

\begin{enumerate}[]
    \item\textbf{Q1.} How does \coolname{} compare to state-of-the-art VLAs on zero-shot manipulation tasks?
    \item\textbf{Q2.}  How important are affordances to policy success?
    \item\textbf{Q3.}  How does \coolname{} compare to existing scene-based policies on room-level queries involving navigation?
    \item\textbf{Q4.} When and why does \coolname{} fail?
\end{enumerate}

\subsection{Tabletop Manipulation}

\textbf{Queries.} We focus on \uline{zero-shot manipulation} where no demonstration data is available for fine-tuning. We test 15 queries that are comparable to previous works~\cite{pi0-experiment-wild, gao2026taxonomy} and focus on real-world short-horizon tasks that emphasize spatial reasoning, affordances, and more generally semantic generalization, including multi-object referencing~\cite{ gao2026taxonomy}. Queries in this section require interacting with 14 different objects.

\textbf{Baselines.} To answer \textbf{Q1} and \textbf{Q2}, we consider two baselines and an ablation:
\begin{itemize}
    \item \pifast~\cite{pertsch2025fast}: An autoregressive VLA that can perform a wide-range of manipulation tasks in new scenes. We run \pifast for a maximum of 800 timesteps per query.
    \item \pihalf~\cite{intelligence2025pihalf}: A flow-matching VLA co-trained on heterogeneous data with better open-world generalization. Given the faster inference, we run \pihalf for a maximum of 1,500 timesteps per query.
    \item \coolname{} (No Aff): an ablative baseline where \affpcd is replaced with the object point cloud \objpcd. All other skills are available. This baseline measures how fine-grained affordance segmentation impacts policy success. 
    We only run this baseline on the 9 queries involving affordances and otherwise report the \coolname{} numbers. 
\end{itemize}

For \pifast and \pihalf, we use the \texttt{openpi} checkpoints fine-tuned on DROID~\cite{khazatsky2024droid}. We use a Franka arm and the DROID setup to minimize the risk of any distribution shift with the XArm6 (not in DROID). While $\pi_0$ does not leverage depth data, it does use two third-person cameras in addition to the wrist camera. Overall, we expect both arms to be equally capable on the considered tasks.

The \coolname{} agents may reattempt some skills on failure due to tool feedback. We cap the number of retries at three.

\textbf{Results.} We show some queries in Fig.~\ref{fig:cool_demos} and report manipulation success rates in Fig.~\ref{fig:tabletop_results}. \coolname{} outperforms \pifast and \pihalf on 13 of the 15 queries, showing strong language understanding and manipulation skills. The $\pi$ models perform well on the drawer opening tasks but fail on comparatively simpler picking tasks showing an overall success rate of $\sim$20\%. This is in line with a recent study of \pifast in the zero-shot setting~\cite{pi0-experiment-wild}. We did observe increased success with \pihalf on the duckie queries and the more complex thumbtack query. In a high number of cases, we find that \pifast (67\%) and \pihalf (75\%) show some level of progression towards the task.  (Fig.~\ref{fig:progress_rate_results} and Fig.~\ref{fig:sankey_pi}).

Our ablative baseline shows that affordance detection plays a key role when the tasks go beyond pick-and-place, with \coolname{} outperforming \coolname{} (No Aff) on the keyboard, power adapter, thumbtack, and drawer queries. Affordances also guide manipulation towards more natural handle usage, although some unnatural grasps still occur~(Fig.~\ref{fig:sankey}, top right).

\subsection{Mobile Manipulation}

\textbf{Baselines.} To answer \textbf{Q2} and \textbf{Q3}, we run mobile \coolname{} and the \coolname{} (No Aff) ablative baseline, which this time has access to affordances for manipulation~(\skillswitch) but not navigation~(\navtool): we set $\lambda_{aff}=0$ and skip affordance detection in \navtool. Moreover, we consider an alternative scene-based policy based on \textit{ConceptGraphs} (CG)~\cite{gu2024conceptgraphs}. For CG, we directly prompt a Gemini 2.5 agent with the robot pose and text serializations of $\objsetglob$ and $\objset_t$ following the procedure mentioned in the CG paper~\cite{gu2024conceptgraphs}, which includes object captions. The CG agent is otherwise free to use \navtool and \skillswitch (without \su and affordances).

\textbf{Results.} We describe the mobile queries and present the corresponding results in Fig.~\ref{fig:mobile_results}.  We find that \coolname{} successfully interleaves navigation and manipulation to solve room-level problems, showing how agentic reasoning over $\objsetglob$ can support decision-making beyond the tabletop setting. The double pick queries show some degree of planning proficiency while the spatial queries demonstrate the effectiveness of $\su$ for grounding objects and understanding referential queries. When object affordances face a particular direction, incorporating affordance information at the navigation stage proves critical to query success. CG performs comparably on double pick but drops significantly on affordance and spatial queries. The gap in affordance queries confirms that scene representations without explicit affordances struggle beyond pick-and-place tasks, while the gap in spatial queries highlights that explicit grounding tools $\su$ are more effective than direct LLM prompting over serialized scene graphs.

\subsection{Analysis}
To answer \textbf{Q4}, we explore and discuss the \coolname{} failure modes in Fig.~\ref{fig:sankey}. Perception is behind a number of failures. We also compare the \coolname{} and VLA runtimes in Table~\ref{tab:runtime}.

\section{DISCUSSION}

\textbf{Limitations and Future Work.} The \coolname{} design has been optimized for short-term language-conditioned manipulation tasks.
Despite already having some planning capabilities (see the double picking experiments, Fig.~\ref{fig:mobile_results}), \coolname{} would benefit from improvements on both the scene representation front (dynamic memory~\cite{liu2025dynamem}, hierarchical memory~\cite{hovsg}, exploration) and the task planning front to tackle long-horizon problems. Additional work would also be required to handle severe occlusions and hidden objects in dense environments.

 More fundamentally, affordance-based skill implementations do not provide a clear avenue for solving long-horizon problems with complex motions, such as \texttt{clean the counter}, \texttt{fold the shirt}, or \texttt{make the bed}---tasks that are achievable through imitation learning and VLAs (generalization notwithstanding). While this suggests a promising integration of structured scene representations and low-level learned policies, our results indicate that scene-based policies can effectively generalize to \uline{more objects} than VLAs can currently interact with in a zero-shot setting. Learned skills would increase the ceiling of \coolname{} in terms of manipulation complexity, but would likely hinder semantic generalization. This non-trivial integration presents an interesting problem for future work.

\begin{table}[t]
\centering
\caption{Average Runtime ($s$) over five successful \texttt{open the drawer} queries. The $\pi_i$ column refers to actual skill execution. While \polasp is slower overall due to perception and tool calling overhead, skill execution is faster than the compared methods.}
\label{tab:runtime}
\setlength{\tabcolsep}{1.8pt} 
\scriptsize
\begin{tabular}{@{}lccccc@{}}
\toprule
\textbf{Method} & $\mapfunc$ & $\affmap$ & $\pi_i$ & \textbf{Rest} & \textbf{Total} \\ \midrule
\pifast        & -           & -           & 44.3           & -           & 44.3           \\
\pihalf        & -           & -           & 36.6           & -           & 36.6           \\
\polasp (Ours)       & 4.3            & 18.2           & 16.8           & 16.0           & 55.3           \\
\bottomrule
\end{tabular}
\vspace{-6.0mm}
\end{table}

\section*{ACKNOWLEDGMENTS}

The work at the Université de Montréal was supported by the Natural Sciences and Engineering Research Council of Canada (NSERC) (Paull) and an NSERC PGS D Scholarship (Morin). Moreover, this research was enabled in part by compute resources provided by Mila (mila.quebec). 
This work has been carried out while Francesco Argenziano was enrolled in the Italian National Doctorate on Artificial Intelligence run by Sapienza University of Rome. This research was conducted while Francesco Argenziano was enrolled as a visiting researcher at Mila - Quebec AI Institute.

\bibliographystyle{IEEEtran}
\bibliography{IEEEtranBST/IEEEabrv, root}

\begin{thebibliography}{10}
\providecommand{\url}[1]{#1}
\csname url@samestyle\endcsname
\providecommand{\newblock}{\relax}
\providecommand{\bibinfo}[2]{#2}
\providecommand{\BIBentrySTDinterwordspacing}{\spaceskip=0pt\relax}
\providecommand{\BIBentryALTinterwordstretchfactor}{4}
\providecommand{\BIBentryALTinterwordspacing}{\spaceskip=\fontdimen2\font plus
\BIBentryALTinterwordstretchfactor\fontdimen3\font minus \fontdimen4\font\relax}
\providecommand{\BIBforeignlanguage}[2]{{%
\expandafter\ifx\csname l@#1\endcsname\relax
\typeout{** WARNING: IEEEtran.bst: No hyphenation pattern has been}%
\typeout{** loaded for the language `#1'. Using the pattern for}%
\typeout{** the default language instead.}%
\else
\language=\csname l@#1\endcsname
\fi
#2}}
\providecommand{\BIBdecl}{\relax}
\BIBdecl

\bibitem{zitkovich2023rt2}
B.~Zitkovich \emph{et~al.}, ``Rt-2: Vision-language-action models transfer web knowledge to robotic control,'' in \emph{Conference on Robot Learning}.\hskip 1em plus 0.5em minus 0.4em\relax PMLR, 2023, pp. 2165--2183.

\bibitem{pertsch2025fast}
K.~Pertsch \emph{et~al.}, ``Fast: Efficient action tokenization for vision-language-action models,'' \emph{arXiv preprint arXiv:2501.09747}, 2025.

\bibitem{intelligence2025pihalf}
P.~Intelligence \emph{et~al.}, ``$\pi_{0.5}$: a vision-language-action model with open-world generalization,'' \emph{arXiv preprint arXiv:2504.16054}, 2025.

\bibitem{gao2026taxonomy}
J.~Gao \emph{et~al.}, ``A taxonomy for evaluating generalist robot manipulation policies,'' \emph{IEEE Robotics and Automation Letters}, 2026.

\bibitem{hancock2025actions}
A.~J. Hancock \emph{et~al.}, ``Actions as language: Fine-tuning vlms into vlas without catastrophic forgetting,'' \emph{arXiv preprint arXiv:2509.22195}, 2025.

\bibitem{vlmaps}
C.~Huang, O.~Mees, A.~Zeng, and W.~Burgard, ``Visual language maps for robot navigation,'' \emph{arXiv preprint arXiv:2210.05714}, 2022.

\bibitem{liu2024ok}
P.~Liu \emph{et~al.}, ``Ok-robot: What really matters in integrating open-knowledge models for robotics,'' \emph{arXiv preprint arXiv:2401.12202}, 2024.

\bibitem{gu2024conceptgraphs}
Q.~Gu \emph{et~al.}, ``Conceptgraphs: Open-vocabulary 3d scene graphs for perception and planning,'' in \emph{IEEE International Conference on Robotics and Automation (ICRA)}.\hskip 1em plus 0.5em minus 0.4em\relax IEEE, 2024, pp. 5021--5028.

\bibitem{liu2025dynamem}
P.~Liu \emph{et~al.}, ``Dynamem: Online dynamic spatio-semantic memory for open world mobile manipulation,'' in \emph{2025 IEEE International Conference on Robotics and Automation (ICRA)}.\hskip 1em plus 0.5em minus 0.4em\relax IEEE, 2025, pp. 13\,346--13\,355.

\bibitem{black2024pi_0}
K.~Black \emph{et~al.}, ``$\pi_0$: A vision-language-action flow model for general robot control,'' \emph{arXiv preprint arXiv:2410.24164}, 2024.

\bibitem{kim2025fine}
M.~J. Kim, C.~Finn, and P.~Liang, ``Fine-tuning vision-language-action models: Optimizing speed and success,'' \emph{arXiv preprint arXiv:2502.19645}, 2025.

\bibitem{slam-handbook}
L.~Paull \emph{et~al.}, \emph{Towards Open-World Spatial {AI}}.\hskip 1em plus 0.5em minus 0.4em\relax Cambridge University Press.

\bibitem{kassab2025openlex3d}
C.~Kassab \emph{et~al.}, ``Openlex3d: A new evaluation benchmark for open-vocabulary 3d scene representations,'' \emph{arXiv preprint arXiv:2503.19764}, 2025.

\bibitem{koch2024open3dsg}
S.~Koch \emph{et~al.}, ``Open3dsg: Open-vocabulary 3d scene graphs from point clouds with queryable objects and open-set relationships,'' in \emph{Proceedings of Computer Vision and Pattern Recognition}, June 2024.

\bibitem{conceptfusion}
K.~Jatavallabhula \emph{et~al.}, ``Conceptfusion: Open-set multimodal 3d mapping,'' \emph{Robotics: Science and Systems (RSS)}, 2023.

\bibitem{peng2023openscene}
S.~Peng \emph{et~al.}, ``Openscene: 3d scene understanding with open vocabularies,'' in \emph{Proceedings of the IEEE/CVF conference on computer vision and pattern recognition}, 2023, pp. 815--824.

\bibitem{hovsg}
A.~Werby \emph{et~al.}, ``Hierarchical open-vocabulary 3d scene graphs for language-grounded robot navigation,'' in \emph{First Workshop on Vision-Language Models for Navigation and Manipulation at ICRA 2024}, 2024.

\bibitem{lu2023ovir}
S.~Lu \emph{et~al.}, ``Ovir-3d: Open-vocabulary 3d instance retrieval without training on 3d data,'' in \emph{Conference on Robot Learning}.\hskip 1em plus 0.5em minus 0.4em\relax PMLR, 2023, pp. 1610--1620.

\bibitem{delitzas2024scenefun3d}
A.~Delitzas \emph{et~al.}, ``Scenefun3d: Fine-grained functionality and affordance understanding in 3d scenes,'' in \emph{Proceedings of Computer Vision and Pattern Recognition}, 2024, pp. 14\,531--14\,542.

\bibitem{Winston1983LearningPD}
\BIBentryALTinterwordspacing
P.~H. Winston, B.~Katz, T.~O. Binford, and M.~R. Lowry, ``Learning physical descriptions from functional definitions, examples, and precedents,'' in \emph{AAAI Conference on Artificial Intelligence}, 1983. [Online]. Available: \url{https://api.semanticscholar.org/CorpusID:7856739}
\BIBentrySTDinterwordspacing

\bibitem{affordancenet}
T.-T. Do, A.~Nguyen, and I.~Reid, ``Affordancenet: An end-to-end deep learning approach for object affordance detection,'' in \emph{2018 IEEE International Conference on Robotics and Automation (ICRA)}, 2018, pp. 5882--5889.

\bibitem{Luddecke_2017_ICCV}
T.~Luddecke and F.~Worgotter, ``Learning to segment affordances,'' in \emph{Proceedings of the IEEE International Conference on Computer Vision (ICCV) Workshops}, Oct 2017.

\bibitem{Nagarajan2020LearningAL}
\BIBentryALTinterwordspacing
T.~Nagarajan and K.~Grauman, ``Learning affordance landscapes for interaction exploration in 3d environments,'' \emph{ArXiv}, vol. abs/2008.09241, 2020. [Online]. Available: \url{https://api.semanticscholar.org/CorpusID:221246369}
\BIBentrySTDinterwordspacing

\bibitem{openfungraph}
C.~Zhang \emph{et~al.}, ``Open-vocabulary functional 3d scene graphs for real-world indoor spaces,'' \emph{Proceedings of Computer Vision and Pattern Recognition}, pp. 19\,401--19\,413, 2025.

\bibitem{fun3du}
J.~Corsetti \emph{et~al.}, ``Functionality understanding and segmentation in 3d scenes,'' \emph{Proceedings of Computer Vision and Pattern Recognition}, pp. 24\,550--24\,559, 2024.

\bibitem{tong2024oval}
E.~Tong \emph{et~al.}, ``Oval-prompt: Open-vocabulary affordance localization for robot manipulation through llm affordance-grounding,'' \emph{arXiv preprint arXiv:2404.11000}, 2024.

\bibitem{qian2024affordancellm}
S.~Qian \emph{et~al.}, ``Affordancellm: Grounding affordance from vision language models,'' in \emph{Proceedings of the IEEE/CVF Conference on Computer Vision and Pattern Recognition}, 2024, pp. 7587--7597.

\bibitem{team2023gemini}
G.~Team \emph{et~al.}, ``Gemini: a family of highly capable multimodal models,'' \emph{arXiv preprint arXiv:2312.11805}, 2023.

\bibitem{code-as-policies}
J.~Liang \emph{et~al.}, ``Code as policies: Language model programs for embodied control,'' \emph{arXiv preprint arXiv:2209.07753}, 2022.

\bibitem{singh2022progprompt}
I.~Singh \emph{et~al.}, ``Progprompt: Generating situated robot task plans using large language models,'' \emph{arXiv preprint arXiv:2209.11302}, 2022.

\bibitem{saycan}
M.~Ahn \emph{et~al.}, ``Do as i can, not as i say: Grounding language in robotic affordances,'' \emph{arXiv preprint arXiv:2204.01691}, 2022.

\bibitem{voxposer}
W.~Huang \emph{et~al.}, ``Voxposer: Composable 3d value maps for robotic manipulation with language models,'' \emph{arXiv preprint arXiv:2307.05973}, 2023.

\bibitem{yao2023react}
S.~Yao \emph{et~al.}, ``React: Synergizing reasoning and acting in language models,'' in \emph{International Conference on Learning Representations (ICLR)}, 2023.

\bibitem{robot-agent-survey}
S.~Salimpour \emph{et~al.}, ``Towards embodied agentic ai: Review and classification of llm-and vlm-driven robot autonomy and interaction,'' \emph{arXiv preprint arXiv:2508.05294}, 2025.

\bibitem{llmgrounder}
J.~Yang \emph{et~al.}, ``Llm-grounder: Open-vocabulary 3d visual grounding with large language model as an agent,'' in \emph{2024 IEEE International Conference on Robotics and Automation (ICRA)}.\hskip 1em plus 0.5em minus 0.4em\relax IEEE, 2024, pp. 7694--7701.

\bibitem{rosa}
R.~Royce \emph{et~al.}, ``Enabling novel mission operations and interactions with rosa: The robot operating system agent,'' in \emph{2025 IEEE Aerospace Conference}.\hskip 1em plus 0.5em minus 0.4em\relax IEEE, 2025, pp. 1--16.

\bibitem{fang2023anygrasp}
H.-S. Fang \emph{et~al.}, ``Anygrasp: Robust and efficient grasp perception in spatial and temporal domains,'' \emph{IEEE Transactions on Robotics}, vol.~39, no.~5, pp. 3929--3945, 2023.

\bibitem{guo2026agenticlab}
P.~Guo \emph{et~al.}, ``Agenticlab: A real-world robot agent platform that can see, think, and act,'' \emph{arXiv preprint arXiv:2602.01662}, 2026.

\bibitem{chen2026gap}
K.~Chen \emph{et~al.}, ``Gap: A graph-as-policy multi-agent self-learning harness for variational automation tasks,'' \emph{arXiv preprint arXiv:2607.05369}, 2026.

\bibitem{coala}
T.~Sumers, S.~Yao, K.~R. Narasimhan, and T.~L. Griffiths, ``Cognitive architectures for language agents,'' \emph{Transactions on Machine Learning Research}, 2023.

\bibitem{clip}
A.~Radford \emph{et~al.}, ``Learning transferable visual models from natural language supervision,'' in \emph{International conference on machine learning}.\hskip 1em plus 0.5em minus 0.4em\relax PmLR, 2021, pp. 8748--8763.

\bibitem{mobilesam}
C.~Zhang \emph{et~al.}, ``Faster segment anything: Towards lightweight sam for mobile applications,'' \emph{arXiv preprint arXiv:2306.14289}, 2023.

\bibitem{sam}
A.~Kirillov \emph{et~al.}, ``Segment anything,'' in \emph{Proceedings of the IEEE/CVF international conference on computer vision}, 2023, pp. 4015--4026.

\bibitem{xiong2024adaptive}
H.~Xiong, R.~Mendonca, K.~Shaw, and D.~Pathak, ``Adaptive mobile manipulation for articulated objects in the open world,'' \emph{arXiv preprint arXiv:2401.14403}, 2024.

\bibitem{chitta2012moveit}
S.~Chitta, I.~Sucan, and S.~Cousins, ``Moveit![ros topics],'' \emph{IEEE robotics \& automation magazine}, vol.~19, no.~1, pp. 18--19, 2012.

\bibitem{rtabmap}
M.~Labb{\'e} and F.~Michaud, ``Rtab-map as an open-source lidar and visual simultaneous localization and mapping library for large-scale and long-term online operation,'' \emph{Journal of Field Robotics}, vol.~36, no.~2, pp. 416--446, 2019.

\bibitem{nav2}
S.~Macenski, F.~Martin, R.~White, and J.~Ginés~Clavero, ``The marathon 2: A navigation system,'' in \emph{2020 IEEE/RSJ International Conference on Intelligent Robots and Systems (IROS)}, 2020.

\bibitem{pi0-experiment-wild}
J.~Wang \emph{et~al.}, ``Evaluating pi0 in the wild: Strengths, problems, and the future of generalist robot policies,'' 2025.

\bibitem{khazatsky2024droid}
A.~Khazatsky \emph{et~al.}, ``Droid: A large-scale in-the-wild robot manipulation dataset,'' \emph{arXiv preprint arXiv:2403.12945}, 2024.

\end{thebibliography}
\section*{APPENDIX}

\subsection{Contribution Statement}
\textbf{Sacha Morin} conceptualized and coordinated the project, implementing most of the \coolname{} perception and robot code and writing the majority of the manuscript.

\textbf{Kumaraditya Gupta} designed and implemented the affordance detection pipeline and wrote sections of the manuscript, also designing most figures, and helping with the \coolname{} experiments. Kumaraditya was also pivotal in some early exploratory work, including mobile manipulation in Isaac Sim and segmentation in SceneFun3D.

\textbf{Mahtab Sandhu} led the \pifast and \pihalf experiments, providing a critical comparison with VLAs.

\textbf{Charlie Gauthier} and \textbf{Francesco Argenziano} participated in multiple brainstorming sessions, also respectively contributing to some skill implementations and the deployment of AnyGrasp.

\textbf{Kirsty Ellis} spearheaded the initial integration of the mobile manipulator and provided invaluable hardware support throughout the project.

\textbf{Liam Paull} was the lead advisor on this project, providing critical feedback that shaped the manuscript and the experiments, also writing and proof-reading sections of the manuscript.

\subsection{Progress Rate and VLA Failure Analysis}
\label{subsec:progress_rate}
\begin{figure*}
    \centering
    \includegraphics[width=\textwidth]{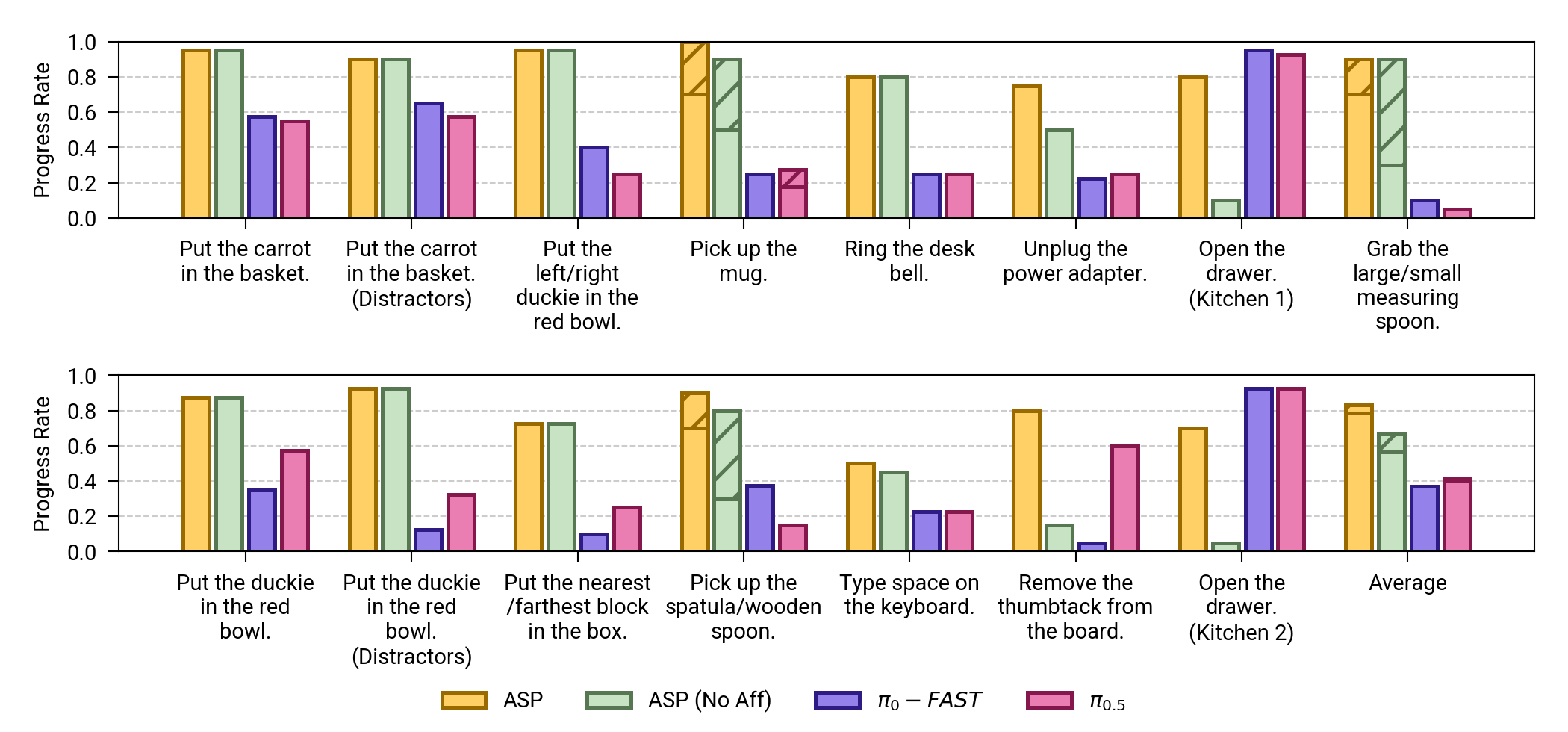}
    \caption{\textbf{Progress Rate.} To complement the success rates in Table~\ref{fig:tabletop_results} we report the average progress rate for \coolname{}, \coolname{} (No Aff), \pifast and \pihalf. We report the average progress rate over 10 attempts per task, giving partial points when the gripper comes close to the relevant object (0.25), when a pick is successful but subsequent steps fail (0.5), and when the policy attempts a reasonable motion (pushing, pulling) on the object without achieving a successful outcome (0.5). Task successes receive a progress rate of 1.00. While the VLAs show some level of task progression on most tasks, they still underperform \coolname{}.}
    \label{fig:progress_rate_results}
\end{figure*}

\begin{figure*}
    \centering
    \includegraphics[width=1.00\textwidth,trim={0 0 0 0},clip]{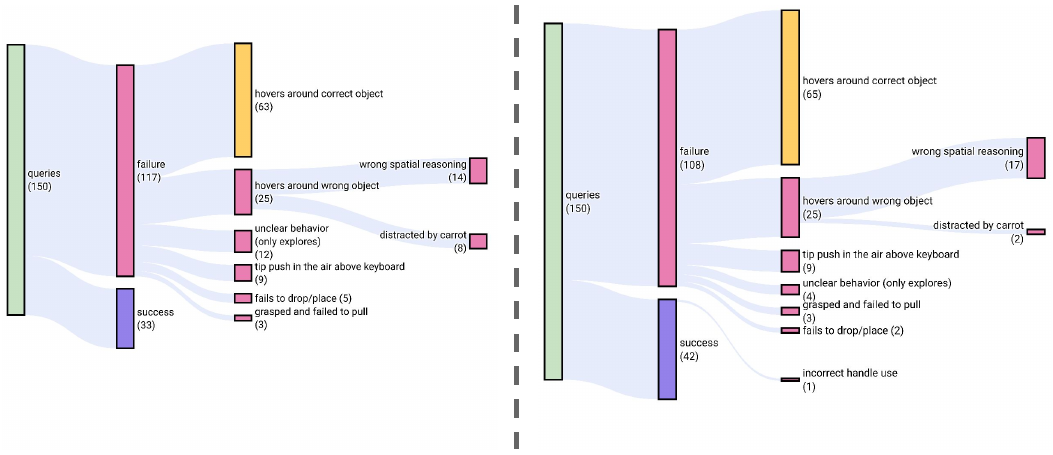}
    \caption{\textbf{Failure Analysis of \pifast and \pihalf.} While the end-to-end nature of \pifast and \pihalf makes it difficult to find a causal explanation for failures, we can still attempt to classify trials in terms of their outcomes and the behaviors of the methods according to a human evaluator. We analyze outcomes for \pifast \textbf{(Left)} and \pihalf \textbf{(Right)}. By far, the most common failure modes involve gripper interactions. Examples include ``hovering" around the correct object without successfully grasping, or partially executing a skill midair above the object (keyboard), suggesting high-level query understanding, but potential issues with depth perception or excessive collision avoidance behaviors. 
    In terms of language following, we found the VLAs to generally struggle with spatial reasoning (\texttt{left/right}, \texttt{near/far}, \texttt{small/large}). \pifast was especially susceptible to distraction by the toy carrot.
    }
    \label{fig:sankey_pi}
\end{figure*}

We study task progression for the manipulation tasks in  Fig.~\ref{fig:progress_rate_results} and some specific \pifast and \pihalf failures in Fig.~\ref{fig:sankey_pi}.

\end{document}